\documentclass[10pt,twocolumn,letterpaper]{article}

\usepackage[pagenumbers]{cvpr} 

\usepackage[dvipsnames]{xcolor}

\definecolor{cvprblue}{rgb}{0.21,0.49,0.74}
\usepackage[pagebackref,breaklinks,colorlinks,citecolor=cvprblue]{hyperref}

\usepackage{amsmath,amsfonts,bm}

\def\eqref#1{equation~\ref{#1}}

\def\1{\bm{1}}

\def\rmX{{\mathbf{X}}}

\DeclareMathAlphabet{\mathsfit}{\encodingdefault}{\sfdefault}{m}{sl}
\SetMathAlphabet{\mathsfit}{bold}{\encodingdefault}{\sfdefault}{bx}{n}

\usepackage{url}
\usepackage{graphicx}
\usepackage{amssymb}
\usepackage{booktabs}

\usepackage{caption}
\usepackage{xcolor}
\usepackage{subcaption}
\usepackage{color,colortbl}

\usepackage{pifont}
\usepackage{adjustbox}    
\usepackage{multirow}
\usepackage{float}
\usepackage{cuted}
\usepackage{capt-of}
\usepackage{tabu}
\usepackage[toc,page,header]{appendix}
\usepackage{minitoc}

\usepackage{xspace}
\usepackage{xcolor,colortbl}
\usepackage{amsmath}
\usepackage[pagebackref,breaklinks,colorlinks]{hyperref}
\usepackage[capitalize]{cleveref}

\usepackage[ruled,lined]{algorithm2e}
\usepackage{setspace}

\usepackage{wrapfig}
\usepackage{makecell}

\usepackage[accsupp]{axessibility}

\makeatletter
\DeclareRobustCommand\onedot{\futurelet\@let@token\@onedot}
\def\@onedot{\ifx\@let@token.\else.\null\fi\xspace}

\newcommand{\cmark}{\ding{51}}%
\newcommand{\xmark}{\ding{55}}%
\makeatother
\usepackage{xcolor,colortbl}

\definecolor{Gray}{gray}{0.85}
\definecolor{LightCyan}{rgb}{0.88,1,1}

\newcolumntype{a}{>{\columncolor{Gray}}c}

\newcommand\clearrow{\global\let\rowmac\relax}
\clearrow

\def\paperID{14} 
\def\confName{CVPR}
\def\confYear{2024}

\title{Federated Learning with a Single Shared Image}

\author{Sunny Soni\\
\small University of Amsterdam\\
{\tt\small sunnysoni97@gmail.com}
\and
Aaqib Saeed\\
\small TU Eindhoven\\
{\tt\small a.saeed@tue.nl}
\and
Yuki M. Asano \\
\small University of  Amsterdam\\
{\tt \small y.m.asano@uva.nl}
}

\usepackage[symbol]{footmisc}

\begin{document}
\maketitle
\doparttoc
\faketableofcontents

\begin{figure*}[htbp]
    \centering
    \includegraphics[width=1.0\textwidth]{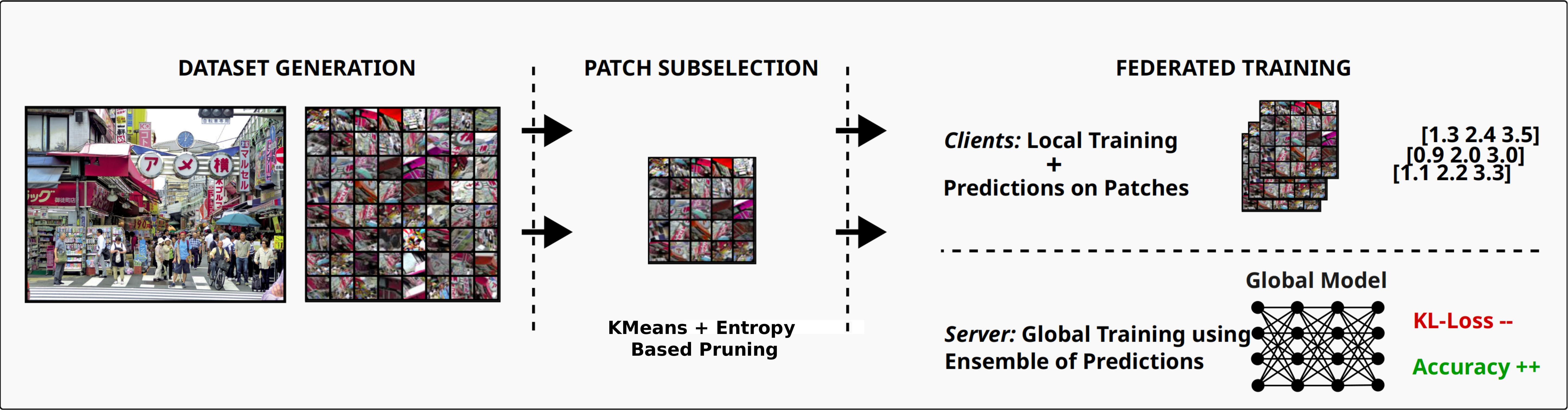}
    \captionof{figure}{Illustration of our federated learning algorithm using a single image. Our algorithm works on the principle of generating a common distillation dataset from only one shared single image using deterministic augmentations. To this end, our method dynamically selects the best patches for the training of the global model in the next round using knowledge distillation. 
    }
    \label{fig:experiments:setup}
\end{figure*}

\begin{abstract}
Federated Learning (FL) enables multiple machines to collaboratively train a machine learning model without sharing of private training data. 
Yet, especially for heterogeneous models, a key bottleneck remains the transfer of knowledge gained from each client model with the server.
One popular method, FedDF, uses distillation to tackle this task with the use of a common, shared dataset on which predictions are exchanged.
However, in many contexts such a dataset might be difficult to acquire due to privacy and the clients might not allow for storage of a large shared dataset. 
To this end, in this paper, we introduce a new method that improves this knowledge distillation method to only rely on a single shared image between clients and server. 
In particular, we propose a novel adaptive dataset pruning algorithm that selects the most informative crops generated from only a single image. 
With this, we show that federated learning with distillation under a limited shared dataset budget works better by using a single image compared to multiple individual ones. 
Finally, we extend our approach to allow for training heterogeneous client architectures by incorporating a non-uniform distillation schedule and client-model mirroring on the server side.
\end{abstract}    
\section{Introduction}\label{s:intro}

Federated Learning (FL) is a paradigm in the field of distributed machine learning that enables multiple clients to collaboratively train powerful predictive models without the need to centralize the training data \citep{zhang2021survey}. It comes with its own set of key challenges in terms of skewed non-IID distribution of data between the participating clients \citep{zhu2021federated, li2020federated, chai2019towards, hsu2019measuring, lin2020ensemble} and communication efficiency during training \citep{konevcny2016federated, lin2020ensemble} among others. These challenges are not directly answered with the classical approaches such as FedAvg \citep{mcmahan2017communication}, which rely primarily on a naive client network parameter sharing approach. Since the inclusion of clients with different data distributions has a factor of heterogeneity involved \citep{zhu2021federated, hsu2019measuring}, another well-known work \citep{li2020federated} counteracts this heterogeneity directly during the client training. This tries to solve one challenge related to non-iidness in private data distribution, but other key challenges related to network parameter sharing remain including concerns with privacy leakage during parameter sharing \citep{wang2019beyond, sun2021soteria}, heterogeneity of client architectures \citep{lin2020ensemble, chai2019towards} and high bandwidth cost of parameter sharing \citep{konevcny2016federated}. To this end, along a second line of thought implementing a server-side training regime, approaches suggested in \citep{lin2020ensemble, li2019fedmd, zhu2021data, sui-etal-2020-feded} make use of the process of knowledge distillation (KD) \citep{hinton2015distilling, gou2021knowledge} to overcome these challenges without the exclusive need of network parameter sharing. To facilitate central network training with the help of KD, the sharing of public data is needed between the clients and the server. 

In this work, we propose a novel approach of making use of a single datum source to act as the shared distillation dataset in ensembled distillation-based federated learning strategies. Our approach makes use of a novel adaptive dataset pruning algorithm on top of generating the distillation data from a single source image during the training. This combination of shared data generation and instance selection process not only allows us to train the central model effectively but also outperforms the other approaches which make use of multiple small-sized images in place of a single image under a limited shared dataset budget. The use of a single datum source has added benefits in domains, where publicly available data and client resources (e.g., network bandwidth and connectivity) are limited in nature. The use of a single datum source has been explored \citep{asano2019critical, asano2022augmented} under the settings of self-supervised learning and understanding extrapolation capabilities of neural networks with knowledge distillation, but it has not yet been explored in federated setting for facilitating model training.

We perform a series of experiments to examine the viability of our proposed algorithm under varying conditions of heterogeneity in private client data, client-server model architectures, rate of pre-training network initializations before distillation, shared dataset storage budget, and real-world domain of the single images. We also extend our experiments to a use case of heterogeneous client architectures involved during a single federated training with the help of client-model mirroring on the server side. To facilitate this, we keep one copy of the client model of each type on the server end, which acts as a global model for the clients that have the same network architecture. The global models are improved with knowledge distillation after each round of local client training with the help of shared logits over the single image dataset. The results we obtain during the aforementioned experiments demonstrate positive reinforcement towards reaching our goal of efficient federated training using Knowledge Distillation under a limited shared dataset budget.

The primary contributions of this work are :
\begin{enumerate}
    \item Demonstrating the efficiency of a single image as a powerful medium for knowledge transfer in a federated learning setting using knowledge distillation.
    \item Novel algorithm for dynamic data pruning which evolves with the current global model during federated learning.
    \item Extensive evaluation of our proposed methods under a variety of conditions in a federated setting.
\end{enumerate}
\section{Related Work}
\label{s:related_work}

\noindent \textbf{Federated learning with knowledge distillation.}\label{s:related_work:kd}
Knowledge Distillation (KD) \citep{hinton2015distilling} has been shown to successfully transfer the knowledge of an ensemble of neural networks into a single network with the means of output logits over a shared dataset. KD has also been leveraged in federated settings, such as Federated Distillation Fusion (FedDF) \citep{lin2020ensemble} and Federated Ensemble Distillation (FedED) \citep{sui-etal-2020-feded}, where the respective authors make use of KD to allow robust and faster convergence on top of using other ensembling methods such as the ones suggested in Federated Averaging \citep{mcmahan2017communication} for initializing the central network before the distillation training of the global model. On the other hand, authors of works such as Federated Model Distillation (FedMD) \citep{li2019fedmd} have also successfully shown that KD can be used for knowledge transfer in a federated setting for the purpose of client model personalization. However, the application of algorithms such as FedMD is targetted for personalization by client-side knowledge distillation rather than improvement of a central model, hence we have not delved into it in the scope of our research. In the case of ensembling methods, it has been shown in \citep{lin2020ensemble} that in the absence of an ensemble of local parameters before distillation training, the final test performance of the central network tends to suffer. As a result, these methods have been shown by the authors to significantly rely on parameter exchange every round similar to naive parameter exchange-based algorithms such as FedAvg \citep{mcmahan2017communication} for robust performance on top of KD. Since the aforementioned KD-based federated algorithms also require significant regular ensembling using network parameter exchange, our approach focuses on improving this aspect by relying significantly on knowledge distillation with the help of data pruning and augmentations on the shared public dataset, which has not yet been explored in these works.

\noindent \textbf{Communication-efficient federated learning.}\label{s:related_work:efficient_fl}
To solve the high bandwidth costs related to parameter sharing, authors of \citep{caldas2018expanding, konevcny2016federated} have shown that quantization of network parameters before the transfer can significantly reduce the bandwidth costs incurred during their transfer. However, with the application of the same low-bit quantization methods with the KD-based federated learning methods in \citep{lin2020ensemble}, the authors have also shown a significant decrease in the overall test performance of models compared to their non-quantized counterparts. On the other hand to not rely on public data sources, authors of the work \citep{zhu2021data} have successfully shown that data-free approaches using a centrally trained generative network for producing the public shared dataset work robustly. However, this also requires an additional exchange of the generative network parameters before each round, which leads to an increase in the network bandwidth usage itself. In pursuit of reducing the bandwidth costs pertaining to network parameter exchanges as well as shared dataset sharing, these works have not yet made an attempt to make use of a storage-efficient single data source, which can simultaneously generate a public distillation dataset alongside being used for dynamic selection without added bandwidth costs. We explore this in our work.

\noindent \textbf{Single image representation learning.} \label{s:related_work:single_img}
Asano et al. \citep{asano2019critical} have successfully made use of a single image to produce augmented patches for facilitating self-supervised learning of useful representations required for solving various downstream tasks. However, the focus of our work is not on the process of solving tasks with the help of self-supervised learning, but on the implications of making use of the single image patches in a federated setting as a medium of knowledge transfer for training robust classifiers. To this end, in a closely resembling work to our target task, the authors in \citep{asano2022augmented} have shown to be able to successfully use KD with single image patches to transfer the knowledge between a pre-trained network and an untrained network to solve the classification task of ImageNet-1k. However, the experiments by the authors were all based in a non-federated setting. In our work, we explore the usage of single image patches in a federated setting as the shared public distillation dataset and its implications in limited shared dataset budget settings.

\section{Methodology}\label{s:method}

Our methods focus on a dynamic procedure to utilize a single image to act as a proxy dataset for distillation in addition to a federated learning setup which is similar to existing ensemble-based knowledge distillation-based methods such as FedDF \citep{lin2020ensemble}. Alongside the generation of a distillation dataset from a single data source, we take care of dynamically selecting the best patches every round to improve the training. The two important parts of our federated strategy are: \textit{Distillation Dataset Generation} (\cref{ss:method:gen}) and \textit{Distillation Dataset Pruning} (\cref{ss:method:select}).
 
\subsection{Distillation Dataset Generation - Patchification}\label{ss:method:gen}

For generating meaningful image representations out of a single image, we make use of the \emph{patchification} technique. Using this technique, we generate a large number of small-size crops from a single image by making combined use of augmentations such as 35-degree rotations, random horizontal flipping, color transformations, etc. We use the same order of augmentations as the one tested in \citep{asano2022augmented} for knowledge distillation-based learning. The image generation procedure is controlled by a given seed to ensure consistent dataset generation when required. This method is communication bandwidth efficient because: 
\begin{enumerate}
    \item It requires the transfer of a single image only once across the network to all the clients during the entire training, which may also be present before the training itself.
    \item The distillation dataset generation is done locally using a seed number, hence only index values need to be transferred when selecting a subset every round.
\end{enumerate}
The generated dataset is used as the proxy dataset for improving the global model using Knowledge Distillation with clients' current predictions. Due to the flexibility provided by augmentations in combination with the subset selection procedure described in \cref{ss:method:select}, one can make use of a single image to produce a varying desired number of patches for the fixed amount of single image data.

\subsection{Distillation Dataset Pruning - Subset Selection}\label{ss:method:select}

After obtaining an initial dataset for knowledge distillation using the method in \cref{ss:method:gen}, we apply dataset pruning methods to it to ensure the selection of information-rich patches for the current round of federated training. The dataset generation procedure makes use of a single information-rich image to generate the small patches, due to which it can produce bad representation patches such as: containing no entities, overlapping with others, being dissimilar to the target domain, and having similar problems arising due to heavy augmentations and presence of information-less regions of the single image. To prune the bad patches, we make use of the following two mechanisms: \textit{KMeans Based Class Balancing} (\cref{ss:method:select:kmeans}) and \textit{Entropy Based Pruning} (\cref{ss:method:select:entropy}). These mechanisms depend on the current global model for their operation, which makes them dynamic in nature and improves their data-pruning ability with the improvement in the global model. Hence, better global models yield better representations.

\begin{figure}[htbp]
    \centering
    \begin{subfigure}{1.0\linewidth}
        \captionsetup{font=scriptsize}
        \includegraphics[width=\linewidth]{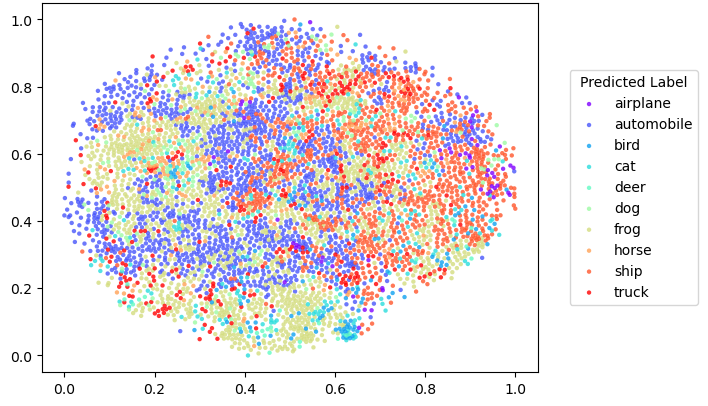}
        \caption{Global Model Accuracy = 52.8}
        \label{fig:img_viz:tsne_low_accuracy}
    \end{subfigure}
    \hfill
    \begin{subfigure}{1.0\linewidth}
        \captionsetup{font=scriptsize}
        \includegraphics[width=\linewidth]{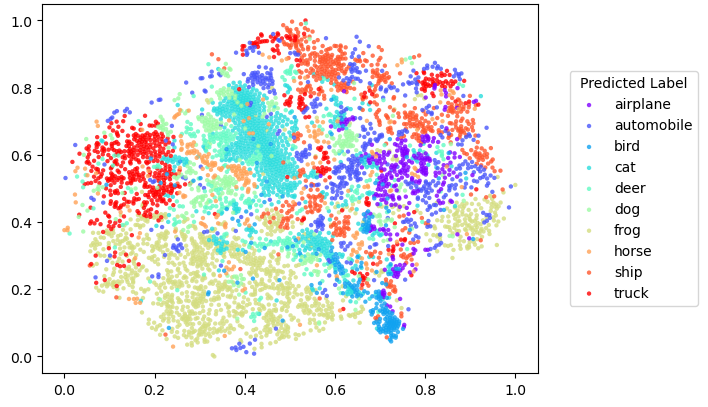}
        \caption{Global Model Accuracy = 76.7}
        \label{fig:img_viz:tsne_high_accuracy}
    \end{subfigure}
    
    \caption{Scatter plot of t-SNE embeddings of pruned single image patches during different phases of training, using our method with FedAvg and ResNet-8 on CIFAR10.}
    
    \label{fig:img_viz:tsne_scatterplot}
\end{figure}

Through the t-SNE visualization in \cref{fig:img_viz:tsne_scatterplot} using a single image with our data pruning methods, we observe the formation of better identifiable boundary structures with a more accurate global model. It provides us with a glimpse of the positive correlation between the effectiveness of our patch subset selection methods with the evolution of the global model.

\noindent \textbf{KMeans-based class balancing.}\label{ss:method:select:kmeans} We first aim to select relevant subcrops that cover the latent space by utilizing K-means. 
Given a distillation dataset $X$, divided in $N$ target classes and its embeddings from the current model $\{f(x_i)\}_{x_i \in X}$, we first generate a KMeans clustering $C(\{f(x_i)\}_{x_i \in X}, K)$ with $K$ clusters, 
\begin{equation}
    C(x_j | \{f(x_i)\}_{x_i \in X}, K) = \min_k(|x_j - c_k|),
\end{equation}
that returns the distance to the nearest centroids $c_k$.
Based on these distances, we establish two cases:
\begin{equation}\label{eqn:kmeans}
H^K(x_j)=
        \begin{cases}
        C(x_j | \{f(x_i)\}_{x_i \in X \cup n}, K) \leq \Theta(n): \text{``Easy"} \\
        C(x_j | \{f(x_i)\}_{x_i \in X \cup n}, K) > \Theta(n):  \text{``Hard"},
    \end{cases}
\end{equation}
where $n \in N$ is the target class for $x_j$ and $\Theta(n)$ is a target class dependent threshold value taken using the following equation:
\begin{equation}
    \begin{split}
    \text{Assuming all }C(x_j | \{f(x_i)\}_{x_i \in X \cup n}, K)\text{ in ascending order,}\\
    \Theta(n) = 
        \begin{cases}
            C(x_j | j = \text{Last Crop to Select}): \text{``Easy"} \\
            C(x_j | j = \text{First Crop to Select}): \text{``Hard"}
        \end{cases}
    \end{split}
\end{equation}

For determining the $j^{\text{th}}$ index in $\Theta(n)$, we depend on the target size of the relevant crop dataset. The full algorithm is supplied in the Appendix A.

\noindent \textbf{Entropy-based pruning.}\label{ss:method:select:entropy} After using K-Means based class balancing, we aim to prune the dataset on the basis of entropy (randomness) present in their output logits using the current global model. This enables us to select samples on the basis of their level of certainty, as predicted by the model to belong to one of the output classes.

Given dataset $X$ divided into $N$ output classes, the entropy of a sample $x_i \in X$ is given by :

\begin{equation}
    E (x_i | g(x_i)) = \max_n(l_n)
\end{equation}
where $g(x_i)$ represents the output logit values for $x_j$ and $l_n$ is one of the logits.

On the basis of our selection criteria $H^E$ (Top / Bottom), we can select the samples with Highest/Lowest entropy values from each class. The full algorithm is supplied in the Appendix A.

\subsection{Federated Learning with Subset Selection using Single Image Dataset}\label{ss:method:fed_single}
Using the mechanism described in \cref{ss:method:gen} and \cref{ss:method:select}, we can perform federated training using a single image according to the steps in Alg. \ref{alg:fed_single}. \textit{FedAvg Initialisation} makes use of naive weighted averaging of parameters on top of KD training for global model updates, which can be toggled with different uniform schedules. We make use of uniform intervals on the basis of defined rates in this work.

{{
\IncMargin{1.0em}
\begin{algorithm*}[htbp]
\setcounter{AlgoLine}{0}
\SetAlCapHSkip{0.7em}
\DontPrintSemicolon
\small

\caption{Federated Learning with Single Image}\label{alg:fed_single}

\vspace{0.5em}

\Begin{
\vspace{0.5em}

\nl Generate distillation dataset $\rmX$ on all clients and the server independently using a single image and fixed seed.\;

\nl Initialise the global model $\theta^G$ on the server. (More than one in case of heterogeneous architectures)\;

\nl \ForAll{rounds of training}{
    \nl Using data pruning, select indices of distillation training examples.\;
    \nl Send the indices and current global model $\theta^G$ to local clients selected for the next round of training.\;
    \nl \ForAll{Local Clients (Selected) in Parallel}{
        \nl Update local model parameters with the received global model.\;
        \nl Do supervised training using the private dataset to improve the local model.\;
        \nl Predict output on selected distillation training examples.\;
        \nl Send predictions to the server. If FedAvg Initialisation is enabled for the current round, also send the local model parameters.\;
    }
    \nl If FedAvg Initialisation is enabled, initialize the global model for KD training using the weighted average of local parameters.\;
    \nl Using KD training on the weighted average of predictions, update the global model.\; 
}
\vspace{0.5em}
}
\end{algorithm*}
\DecMargin{1.0em}
}%
}

\noindent \textbf{Computation overhead.}
Since the clients using our method only do supervised training similar to FedAvg \citep{mcmahan2017communication}, there's no computation overhead on the client side during the training. The compute cost of forward inferencing a small distillation dataset is negligible compared to the supervised training the clients are subjected to.

\section{Experiments}
\label{s:experiments}

\subsection{Experimental Setup}\label{ss:experiments:setup}

\noindent \textbf{Dataset.} We do our experiments across the following publically available datasets: CIFAR10/100 \citep{krizhevsky2009learning} and MedMNIST (PathMNIST) \citep{yang2023medmnist}. For the distribution of private data among the clients from the collective training data, we use a similar strategy to the one suggested in \citep{hsu2019measuring}, which allows us to control the degree of heterogeneity using the parameter $\alpha$ (lower $\alpha$ = higher degree of non-iidness and vice-versa). We use the full test sets corresponding to the private client datasets as evaluation sets for the global model (testing is only server-side). 10\% of the training examples are held as a validation dataset.

For the shared public dataset, we generate patches out of a single image for all the experiments with our method. For the FedDF experiments, we have made use of CIFAR100 training set for CIFAR10 experiments, unless mentioned otherwise. The single images have been visualized in Appendix A alongside the patches and t-SNE visualizations.

\noindent \textbf{Server-client model architecture.} ResNets (trained from scratch) have been used for most of our experiments as the model of choice for the server and clients \citep{he2016deep}. We have also used \textit{heterogeneous network distribuitions} in \cref{ss:experiments:nnarch:hetero}, with clients having varied network architectures training together.  WideResNets have also been used for some of the experiments \citep{zagoruyko2016wide}. Our method should be directly applicable to all kinds of network architectures that have an intermediate embedding layer present. The models have been explicitly defined in the table descriptions for unambiguity.

\noindent \textbf{Hyper-parameter configuration.} The values of the learning rate (local and global)  have been motivated by the experiments described in Appendix D. We use a client learning rate of 0.01 for ResNet and WResNet, while the distillation training learning rate is 0.005. For KMeans Balancing, we use a KMeans model with 1000 clusters, a class balancing factor of 1.0, and the 'Hard' selection heuristic. For Entropy selection, we remove 90\% of the training examples using the 'Top' removal heuristic (Appendix C). For the experiment in Table \ref{tab:abl_strategy}, we do local client training for 10 epochs and server-side distillation for 250 steps, while 40 epochs and 500 distillation steps have been our choice for other experiments unless mentioned otherwise. We prefer to keep the use of FedAvg initializations to 20\% in our experiments unless mentioned otherwise. For all the experiments, we simulate \textbf{20 private clients}, with a selection probability (C) of 0.4 per training round.

\noindent\textbf{FedAvg Initialisation Rate.} FedAvg Initialisation Rate is responsible for controlling the schedule of FedAvg Initialisations (\cref{ss:method:fed_single}), where the percentage decides the round intervals in which we initialise the central network with them before distillation training. For reference, 20\% means using FedAvg Initialisations every 5th round and 100\% means using them every round.

\subsection{Network and Storage Conditions}\label{ss:experiments:communication}

\noindent \textbf{Comparative analysis of single image performance under limited shared dataset budget.}\label{ss:experiments:communication:dataset}

\begin{figure*}[htbp]
    \centering
    \includegraphics[width=0.55\linewidth]{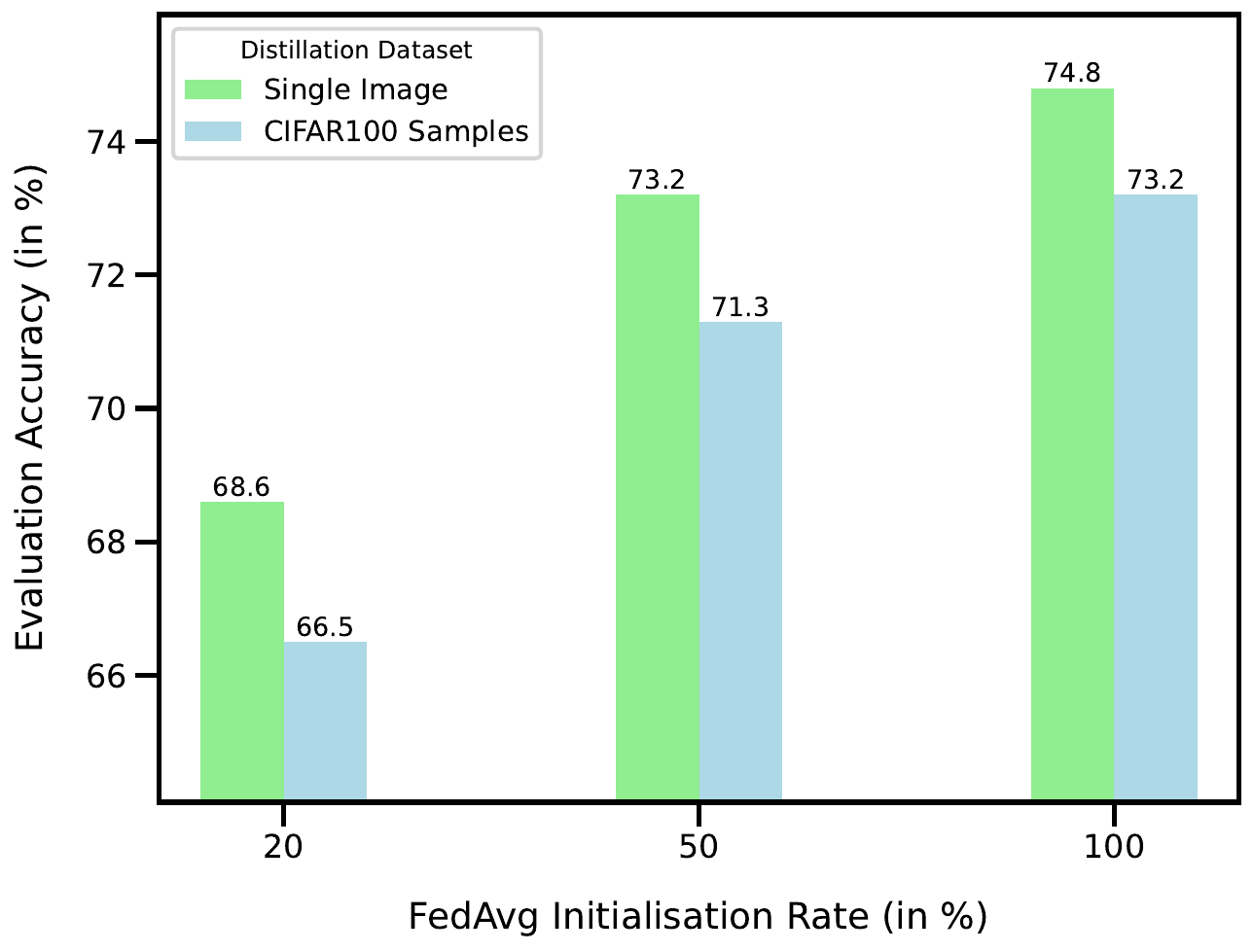}
    \caption{Comparison of test performance in federated setting using a single image with patch selection compared to the equivalent size of multiple independent training samples from a labeled dataset as shared distillation dataset. We use different rates of FedAvg. initializations to emulate different network bandwidth conditions. Detailed result in Table \ref{tab:exp_comm_comparison}.}
    \label{fig:exp_comm_comparison}
\end{figure*}

This is our most significant experiment in terms of exhibiting the viability of federated learning under limited shared dataset budget settings using a single image. Going through the results in Table \ref{tab:exp_comm_comparison}, we see that for the same amount of storage budget, a single image with patch selection outperforms similarly sized individual samples. If we also lower the network budget and the rate of exchange of initialization parameters, it is able to hold at par with individual training samples 10 times its size. This shows a promising future for our work in the scenario where there is limited availability of publicly shared datasets as well as storage budget would be low on participating clients.

{
\begin{table*}[htbp]
\small
    \centering
    \begin{tabular}{lllll}
        \toprule
        \multicolumn{1}{c}{\multirow{2}{*}{\textbf{Distillation Dataset}}} & \multicolumn{1}{c}{\multirow{2}{*}{\textbf{\thead{Number\\of Pixels}}}} & \multicolumn{3}{c}{\textbf{FedAvg Initialisation Rate (in \%)}}\\
        \cmidrule(lr){3-5}
        & & \multicolumn{1}{c}{100} & \multicolumn{1}{c}{50} & \multicolumn{1}{c}{20} \\
        \midrule
        5K CIF100 Samples & 5M & 76.4 $\pm$ 1.4 & 74.1 $\pm$ 1.6 & 68.9 $\pm$ 1.4 \\
        \textbf{Single Image with Patch Selection} & 0.5M & 74.8 $\pm$ 2.6 & 73.2 $\pm$ 3.2 & 68.6 $\pm$ 0.8\\
        500 CIF100 Samples & 0.5M & 73.2 $\pm$ 1.7 & 71.3 $\pm$ 2.0 & 66.5 $\pm$ 0.9 \\
        \bottomrule
    \end{tabular}
    \caption{Best test performance during 30 rounds of training with CIFAR10 Private Data with Distribuition $\alpha$ = 1.0 using ResNet-8 with \textbf{different distillation datasets} and rate of using FedAvg initialisation.}
    \label{tab:exp_comm_comparison}
\end{table*}}

\noindent \textbf{Performance evaluation in limited network bandwidth settings against heterogeneous data distributions.}\label{ss:experiments:communication:init_rate}
 To test the impact of high data distribution heterogeneity on our FL strategy against an existing SOTA federated learning strategy based on knowledge distillation, we show the performance gains in Table \ref{tab:exp_comparison}. We also vary the network initialization rate to test our method in high and low-bandwidth situations. We notice that with the help of patch subset selection, our methods outperform the fed strategy which doesn't make use of this process. This trend is constant across all bandwidth scenarios and local client training expenditures. We have also extended our approach to incorporate FedProx local client training regime, which shows better results than naive local client training. This extendability makes our method unique and viable to more approaches than just one kind of local training which can have added performance benefits with our algorithm.

{\begin{table*}[htbp]
\small
    \centering
    \begin{tabular}{>{\rowmac}l>{\rowmac}l>{\rowmac}l>{\rowmac}l>{\rowmac}l>{\rowmac}l>{\rowmac}l>{\rowmac}l<{\clearrow}}
        \toprule
        \multicolumn{1}{c}{\multirow{3}{*}{\textbf{Strategy}}} & \multicolumn{1}{c}{\multirow{3}{*}{\textbf{\thead{Local\\Epochs}}}} & \multicolumn{6}{c}{\textbf{FedAvg Initialisation Rate (in \%)}} \\
        \cmidrule(lr){3-8}
        & & \multicolumn{2}{c}{100} & \multicolumn{2}{c}{50} & \multicolumn{2}{c}{20} \\
        \cmidrule(lr){3-4} \cmidrule(lr){5-6} \cmidrule(lr){7-8}
        & & \multicolumn{1}{c}{$\alpha=1.0$} & \multicolumn{1}{c}{$\alpha=0.1$} & \multicolumn{1}{c}{$\alpha=1.0$} & \multicolumn{1}{c}{$\alpha=0.1$} & \multicolumn{1}{c}{$\alpha=1.0$} & \multicolumn{1}{c}{$\alpha=0.1$} \\
        \midrule
        \multirow{2}{*}{FedDF} & 20 & 75.7 $\pm$ 1.2 & 48.2 $\pm$ 2.6 & 73.9 $\pm$ 0.8 & 47.3 $\pm$ 5.2 & 71.1 $\pm$ 0.5 & 42.2 $\pm$ 9.4 \\
            & 40 & 75.7 $\pm$ 0.9 & 49.5 $\pm$ 3.1 & 74.9 $\pm$ 1.9 & 49.3 $\pm$ 1.1 & 72.5 $\pm$ 0.5 & 46.1 $\pm$ 6.6\\
        \midrule
        \multirow{2}{*}{Ours w/ FedAvg} & 20 & 76.9 $\pm$ 0.6 & 47.8 $\pm$ 5.3 & 75.8 $\pm$ 0.3 & 47.3 $\pm$ 5.5 & 73.7 $\pm$ 1.0 & 45.5 $\pm$ 5.1 \\
            & 40 & 77.0 $\pm$ 0.6 & 47.8 $\pm$ 5.4 & 76.2 $\pm$ 1.4 & 49.5 $\pm$ 2.2 & 74.3 $\pm$ 0.6 & 46.6 $\pm$ 6.7  \\
        \midrule
        \multirow{2}{*}{\textbf{Ours w/ FedProx}} & 20 & 77.2 $\pm$ 0.8 & 47.2 $\pm$ 7.0 & 74.5 $\pm$ 1.3 & 44.6 $\pm$ 7.9 & 73.1 $\pm$ 0.2 & 46.9 $\pm$ 4.5 \\
            & 40 & 77.7 $\pm$ 0.8 & 47.7 $\pm$ 3.8 & 76.3 $\pm$ 0.4 & 46.0 $\pm$ 5.3 & 74.3 $\pm$ 1.1 & 45.1 $\pm$ 6.0 \\
        \bottomrule
    \end{tabular}%

    \caption{Comparison of best test performance under different settings (FedAvg Initialisation Rate, \textbf{Degree of Heterogenity ($\alpha$)}, Local Training Epochs) using different federated learning strategies with ResNet-8 on CIFAR10 during 30 rounds of training (2 seeds). 5000 single image patches have been used as distillation proxy set (w/o selection mechanism for FedDF).}
    \label{tab:exp_comparison}

\end{table*}}

\subsection{Client-Server Neural Network Architectures}\label{ss:experiments:nnarch}

\noindent \textbf{Evaluating our strategy under homogeneous network Architecture.}\label{ss:experiments:nnarch:homo}
We perform all the experiments in the earlier sections using ResNet-8 as the client and server models. To make sure our federated strategy works equally well among other homogeneous network distributions, we put it to the test against FedDF using ResNet-20 as well as W-ResNet-16-4 in Table \ref{tab:exp_nnarch_homo}. We see that under the same distillation set storage budget, our method works better under all the tested network architectures. As per nominal expectations, network architectures with more parameters show better results than the ones with less number of parameters which enables us to achieve better test performance with more complex networks. Irrespective of the network architecture, the trend is constant when it comes to our FL strategy outperforming other strategies making use of a labelled distillation dataset in a limited storage budget scenario.

{
\begin{table}[htbp]
    \small
    \centering
    \begin{tabular}{lccc}
        \toprule
        \multicolumn{1}{c}{\multirow{2}{*}{\textbf{Fed Strategy}}} & \multicolumn{3}{c}{\textbf{Network Architecture}} \\
        \cmidrule(lr){2-4}
        & ResNet-8 & ResNet-20 & W-ResNet-16-4 \\
        \midrule
        FedDF & 67.3 $\pm$ 1.9 & 73.0 $\pm$ 0.6 & 75.3 $\pm$ 1.2 \\
        \textbf{Ours} & 70.2 $\pm$ 0.8 & 74.1 $\pm$ 0.9 & 75.7 $\pm$ 0.9 \\
        \bottomrule
    \end{tabular}
    \caption{Best test performance during 30 rounds of training using CIF10 Private Data with Distribuition $\alpha = 1.0$ using different Fed strategies and \textbf{homogeneous client-server network architectures} with 20\% rate of FedAvg. initialisation. FedDF uses 500 CIF100 samples as distillation proxy, while our method makes use of a single image of equivalent size with patch subset selection.}
    \label{tab:exp_nnarch_homo}
\end{table}
}

\noindent \textbf{Evaluating our strategy under heterogeneous network architecture.}\label{ss:experiments:nnarch:hetero} In the final experimental section, we test our federated strategy in the presence of heterogeneity in the client model architectures. The results present in Table \ref{tab:exp_nnarch_hetero} show the success of our method in training the global models when pitted against a strategy not utilizing a single image. It also exhibits the importance of constant distillation training for the success of our methods, as our non-uniform approach gives subpar results with less training time. However, when going from 15k to 11k steps, we also save about 1/3 of the training time and computation resources used on the server side. It can be an interesting point of extension to our work to improve upon this non-uniform scheduling to allow for more robust training of heterogeneous models with less computation time.

{

\begin{table}[htbp]
    \small
    \centering
    \begin{tabular}{lcc}
        \toprule
        \multicolumn{1}{c}{\textbf{Fed Strategy}} & \multicolumn{1}{c}{\textbf{\thead{Total\\Distillation Steps}}} & \multicolumn{1}{c}{\textbf{\thead{Macro-Avg Accuracy\\(Hetero Server Models)}}} \\
        \midrule
        FedDF & 15K & 67.4 $\pm$ 0.6  \\
        \textbf{Ours} & 15K & 68.5 $\pm$ 1.1 \\
        \makecell[l]{Ours with\\Scheduling} & 11.3K & 65.2 $\pm$ 1.3 \\
        \bottomrule
    \end{tabular}%
    \caption{Best test performance across  during 30 rounds of training using CIF10 Private Data with Distribuition $\alpha = 1.0$ using different Fed strategies and distillation step scheduling, under a \textbf{heterogenous client architecture distibuition (6 ResNet-8, 7 ResNet-20, 7 W-ResNet-16-4)} with 20\% rate of FedAvg. Initialisation. 500 CIF100 samples have been used as distillation proxy for FedDF, while our method makes use of a Single Image of equivalent size with patch selection.}
    \label{tab:exp_nnarch_hetero}
\end{table}}

\subsection{Patch Selection Mechanism}\label{ss:experiments:patch_selection}
\noindent \textbf{Optimal patch sub selection strategies across private dataset.}\label{ss:experiments:patch_selection:strategy} To find the effectiveness of patch subset selection mechanisms, we test it under different private datasets from different real-world domains (General and Medical). Through Table \ref{tab:abl_strategy}, it is evident that the single image patches work best in the presence of a selection strategy in our federated algorithm. On their own, both KMeans Balancing (\ref{ss:method:select:kmeans}) and Entropy Selection (\ref{ss:method:select:entropy}) strategy works better than employing no selection for the same number of patches. Together, they perform best across all the datasets we have tested which is what we use in our other experiments in this work. Both of the selection strategies and their combination significantly impact the final performance. We have done our primitive analysis with them in light of this work to find an optimal setting (Appendix C, but there might be a correlation between their settings which we have not delved into. We can propose this detailed analysis of their combinative work as future work for improving the test performance of our federated strategy with the means of better data pruning.

{

\begin{table}[htbp]
\small
    \centering
    \resizebox{\linewidth}{!}{%
    \begin{tabular}{lccc}
    \toprule
    \multicolumn{1}{c}{\multirow{2}{*}{\textbf{Selection Strategy}}} & \multicolumn{3}{c}{\textbf{Private Dataset}} \\
    \cmidrule(lr){2-4}
    & \multicolumn{1}{c}{CIFAR10} & \multicolumn{1}{c}{CIFAR100} & \multicolumn{1}{c}{PathMNIST} \\
    \midrule
    No Selection & 63.4 $\pm$ 1.4 & 24.2 $\pm$ 1.1 & 64.5 $\pm$ 4.7 \\
    KMeans & 66.2 $\pm$ 0.8 & 21.8 $\pm$ 2.1 & 67.9 $\pm$ 8.4 \\
    Entropy & 65.9 $\pm$ 1.0 & 26.3 $\pm$ 1.0 & 76.4 $\pm$ 2.8 \\
    \textbf{KMeans + Entropy (Ours)} & 67.0 $\pm$ 1.1 & 26.4 $\pm$ 1.2 & 77.1 $\pm$ 3.0 \\
    \bottomrule
    \end{tabular}%
    }
    \caption{Best test performance achieved during 30 rounds of training with \textbf{different selection mechanisms} (Distillation Set Size = 5000 patches) across different private datasets ($\alpha = 1.0$) using our federated strategy with ResNet-8 while using 20\% rate of FedAvg. initialisations. (2 seeds)}
    \label{tab:abl_strategy}
\end{table}}

\noindent \textbf{Impact of selection mechanism with manually labelled distillation set.}\label{ss:experiments:patch_selection:std_dataset} We test the viability of our selection mechanism in case of extending it to the use cases where we already have a shared public dataset at hand in Table \ref{tab:exp_abl_std}. During the regular exchange of initialization parameters, the application of our selection mechanism exhibits no advantage. However, when we reduce the exchange of initialization parameters to emulate low bandwidth conditions, it shows significant gains. This shows that even with the availability of standard distillation sets at hand in ensembled distillation methods, the subset selection mechanism can play an important role in low bandwidth cost federated training.

{\begin{table}[htbp]
\small
    \centering
    \begin{tabular}{cccc}
        \toprule
        \multicolumn{1}{c}{\multirow{2}{*}{\textbf{\thead{Selection Mechanism \\ Applied}}}} & \multicolumn{3}{c}{\textbf{FedAvg Initialisation Rate (in \%)}}\\
        \cmidrule(lr){2-4}
        & \multicolumn{1}{c}{100} & \multicolumn{1}{c}{50} & \multicolumn{1}{c}{20} \\
        \midrule
        \xmark & 75.0 $\pm$ 0.5  & 73.1 $\pm$ 1.1 & 67.4 $\pm$ 0.5 \\
        \cmark & 73.8 $\pm$ 1.9 & 72.3 $\pm$ 0.6 & 70.7 $\pm$ 1.2 \\
        \bottomrule
    \end{tabular}%
    \caption{Comparison of best test performance during 30 rounds of training with CIFAR10 Private Data with Distribuition $\alpha$ = 1.0 using FedDF (with ResNet-8) between \textbf{use/non-use of selection mechanism} across varying rate of using FedAvg initialisation. 1000 samples from CIFAR100 train split make the distillation proxy dataset.}
    \label{tab:exp_abl_std}
    
\end{table}

}

\subsection{Selecting the Best Source of Single Image}\label{ss:experiments:img_selection}
We conduct cross-dataset single-image testing using our algorithm across 3 private training datasets and 3 images, with two of them corresponding to one of the dataset domains and the third one being a random noise. The results in Table \ref{tab:image_selection} exhibit that it is necessary to use a single image that is similar to the domain of the target task for optimal performance. In the case of using a single random noise image as the distillation proxy, we get the lowest test performance as it is hard for random augmentations to convert random noise patches into a knowledge transfer medium. Hence, care must be taken in choosing a single image with similar patch representations as the target task for optimal learning with our algorithm. There can be an interesting area to explore with more augmentation experiments and generative algorithms if it is possible to use a random noise image viably as a single image with our method. We leave this as future work.

{

\begin{table}[htbp]
    \small
    \centering
    \resizebox{\linewidth}{!}{%
    \begin{tabular}{cccc}
    \toprule
    \multicolumn{1}{c}{\multirow{2}{*}{\textbf{\thead{Private\\Dataset}}}} & \multicolumn{3}{c}{\textbf{Single Image Source}} \\
    \cmidrule(lr){2-4}
    & City Street & Pathology Samples & Random Noise \\
    \midrule
    CIFAR10 & \textbf{75.3} & 69.0 & 39.4 \\
    CIFAR100 & \textbf{32.0} & 12.0 & 6.8 \\
    PathMNIST & 69.7 & \textbf{71.6} & 33.0\\
    \bottomrule
    \end{tabular}%
    }
    \caption{Best test performance during 30 rounds of training using our federated method with \textbf{varying Private Datasets} (Distribution $\alpha = 100.0$) and \textbf{varying Single Image Sources} (5k Patches) (Distillation Proxy Set) on ResNet-8 architecture with 20\% rate of FedAvg. initialization.}
    \label{tab:image_selection}
\end{table}}

\section{Conclusion}
\label{s:conclusion}

Through this work, we present a novel approach for federated learning using ensembled knowledge distillation with the use of augmented image patches from a single image with patch subset selection. We successfully exhibit the performance gains with our approach in a limited shared dataset budget scenario as well as low network bandwidth requiring scenarios with less exchange of network parameters. Alongside low resource usage, the use of a single image also enables our federated strategy to be applicable to scenarios where we have a lack of public datasets required during federated training of multiple clients.\\

\noindent \textbf{Future Work.}\label{ss:conclusion:future} We mention a few specialized avenues of extension to our work during the discussion of results in Section \ref{s:experiments}. Some of the key points that were not mentioned in it include: Application of the single datum-based federated learning to other modalities and machine learning tasks; Application of our work to other knowledge distillation-based algorithms in federated learning other than ensembled methods, such as FedMD \citep{li2019fedmd}; Analysis of different kind of augmentations to improve the robustness of our method. With the aforementioned points, significant work can be done to improve the viability of our novel approach presented in this work to incorporate more real-world challenges.


\bibliographystyle{ieeenat_fullname}
\bibliography{main}

\clearpage
\appendix
\pagenumbering{arabic}
\addcontentsline{toc}{section}{Appendices}
\part{\LARGE{{Federated Learning with a Single Shared Image}}\\
\vspace{2em}
\Large{Appendix}}

{
\hypersetup{linkcolor=black}
\parttoc 
}

\bigskip


\renewcommand{\footnotesize}{\fontsize{7pt}{11pt}\selectfont}

\section{Detailed Algorithm : Dataset Pruning}\label{App:algo_pruning}

\subsection{KMeans Based Class Balancing}\label{App:algo_pruning:kmeans}

Alg. \ref{alg:kmeans_selection} describes the implementation of our KMeans-based balancing strategy used for data pruning.

{{
\IncMargin{1.0em}
\begin{algorithm}[htbp]
\setcounter{AlgoLine}{0}
\SetAlCapHSkip{0.7em}
\DontPrintSemicolon
\small

\caption{K-Means Balancing}\label{alg:kmeans_selection}

\SetKwInOut{Params}{Parameters}

\vspace{0.5em}
\KwIn{Distillation Training Dataset ($X$), Current Global Model ($M^G$)}
\Params{Number of Clusters ($K$), Size of New Dataset ($s$), Balancing Factor ($F^K$), Selection Heuristic ($H^K$)}
\KwOut{Pruned Distillation Training Dataset with KMeans Selection ($X^K$)}

\vspace{0.5em}

\Begin{
\vspace{0.5em}

\nl For all $x_n \in X : n \in [1..S]$ where S = size of X, find $Z = \{z_n : z_n = \text{Embedding Representation }(M^G, x_n)\}$ and $Y = \{y_n : y_n = \text{Max-Index }(\text{Classifier Output }(M^G, x_n)\}$. \;

\nl Define $C^P = \{ \text{Set of unique classes in Y} \}$ and Number of unique classes $C = |C^P|$. \;

\nl Initialise an independent unsupervised KMeans Clustering Model ($M^C$) using $K$ number of cluster centers. Fit $M^C$ on $Z$ and find $D = \{d_n : d_n = $ \{Shortest euclidean distance of $z_n$ to its cluster center\}. \;

\nl Define the minimum number of examples (balancing lower bound) to be selected from each class $c_i \in C^P$, as $LB = [\frac{s}{C} * F^K]$. \;

\nl \ForAll{$c_i \in C^P : i \in [1..C]$}{

    Find the indices of examples belonging to the $c_i$ using $y_n \in Y : y_n = c_i$. \;
    Select indices of the new training examples on the basis of selection heuristic $H^K \in \{\text{Easy, Hard, Mixed}\}$ with their corresponding cluster distance values $D$.\;
    Push the training examples from $X$ with selected indices in the new dataset ($X^K$).\;
    Remove the selected training examples from $X$ and $D$.\;
}

\nl Remaining number of examples to be selected can be calculated as given by : $s - \text{size of }X^K$.\;

\nl Using selection heuristic = $H^K$ on the cluster distance values in $D$, find the indices of the remaining examples to be selected. Push the training examples with selected indices in the new dataset ($X^K$)\;
\vspace{0.5em}
}

\end{algorithm}

\DecMargin{1.0em}
}}

\subsection{Entropy Based Pruning}\label{App:algo_pruning:entropy}

Alg. \ref{alg:entropy_selection} describes the implementation of our Entropy-based pruning strategy used for data pruning.

{{
\IncMargin{1.0em}
\begin{algorithm}[htbp]
\setcounter{AlgoLine}{0}
\SetAlCapHSkip{0.7em}
\DontPrintSemicolon
\small

\caption{Entropy Selection}\label{alg:entropy_selection}

\SetKwInOut{Params}{Parameters}

\vspace{0.5em}
\KwIn{Distillation Training Dataset ($X$), Current Global Model ($M^G$)}
\Params{Percentage of Examples to Prune ($e$), Removal Heuristic ($H^E$)}
\KwOut{Pruned Distillation Training Dataset with Entropy Selection ($X^E$)}

\vspace{0.5em}

\Begin{
\vspace{0.5em}

\nl For all $x_n \in X : n \in [1..S] \text{where S = size of X}$, find $Y = \{ y_n : y_n = \text{Max(Softmax (Classifier Output }(M^G,x_n)))\}$.\;

\nl Select indices of the $(100-e)\%$ training examples using the removal heuristic $H^E \in \{\text{Top, Bottom, Random}\}$ with their corresponding values in $Y$.\;


\nl Push the selected indices in the new dataset ($X^E$).\;

\vspace{0.5em}
}
\end{algorithm}
\DecMargin{1.0em}
}%
}

\section{Image and Patch Visualisations}\label{App:img_viz}

\subsection{Visualisation of Single Images}\label{App:img_viz:images}

\begin{figure*}[!t]
    \centering
    \begin{subfigure}{0.3\textwidth}
        \includegraphics[width=\textwidth]{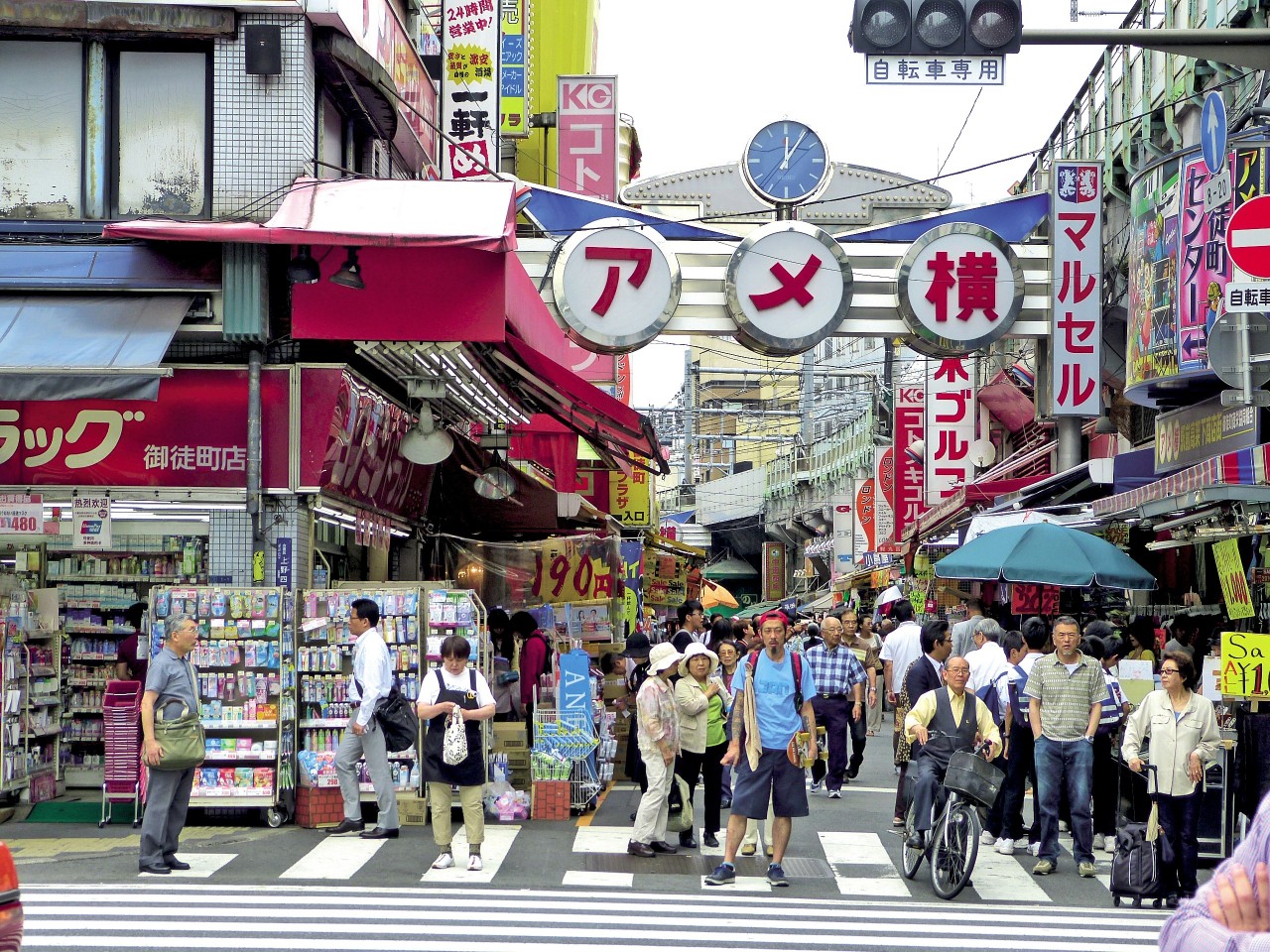}
        \caption{A busy street in a city.}
        \label{fig:img_viz:ameyoko}
    \end{subfigure}
    \begin{subfigure}{0.328\textwidth}
        \includegraphics[width=\textwidth]{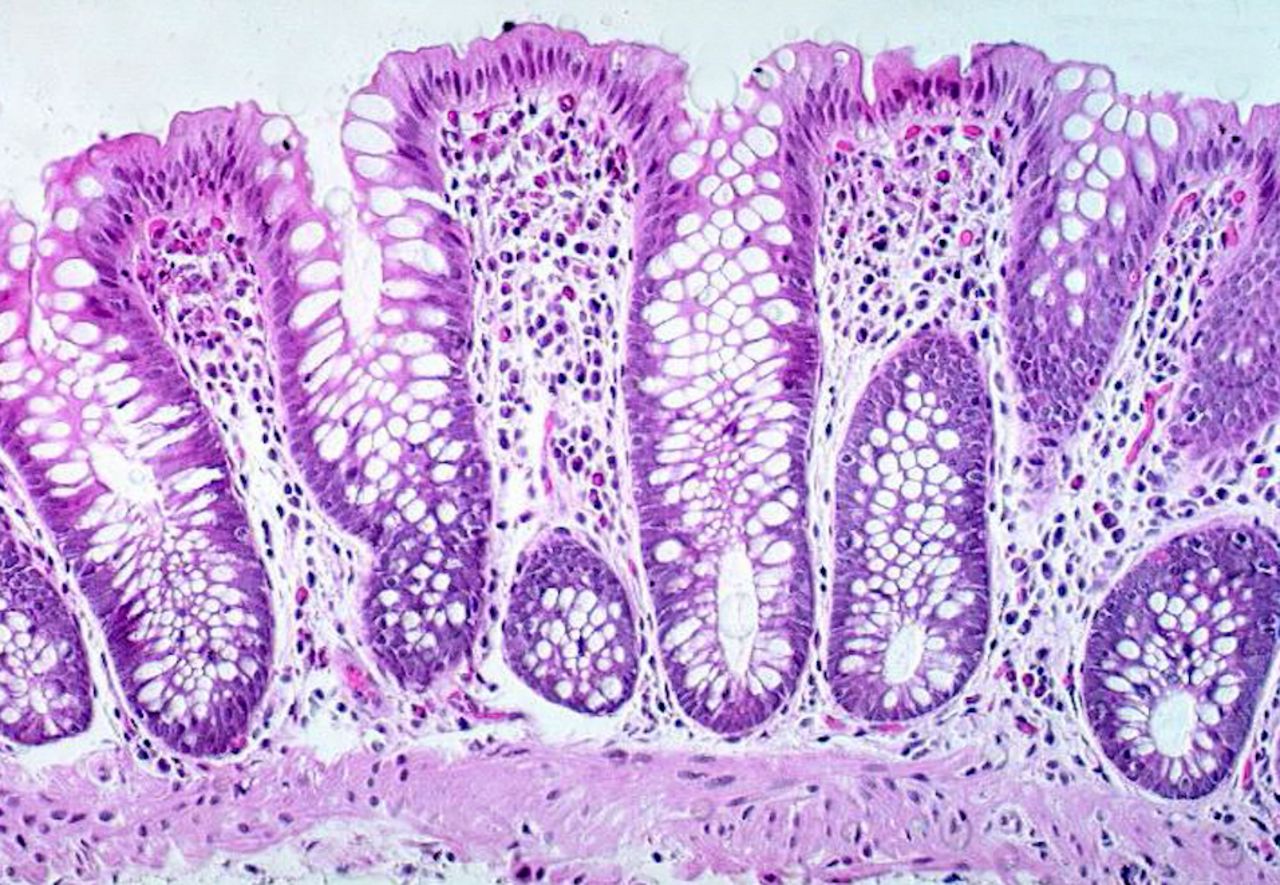}
        \caption{Multiple colon pathology samples.}
        \label{fig:img_viz:colon_pathology}
    \end{subfigure}
    \begin{subfigure}{0.3\textwidth}
        \includegraphics[width=\textwidth]{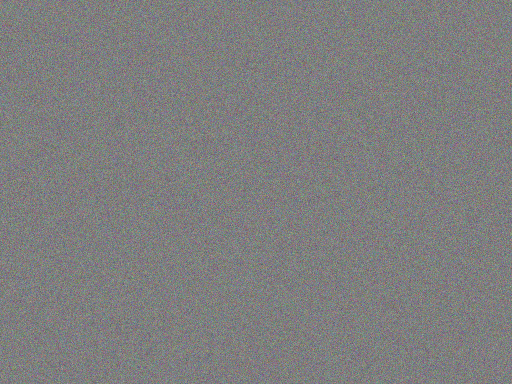}
        \caption{Random noise.}
        \label{fig:img_viz:random_noise}
    \end{subfigure}
    \caption{Single Image sources used for our experiments for distillation dataset generation.}
    \label{fig:img_viz:all_images}
\end{figure*}

We make use of the images depicted in Fig. \ref{fig:img_viz:all_images} as the sources for generating our distillation dataset. Kindly note, that we only used the images for non-profit educational research purposes and we do not hold any rights over their commercial use. These images have been selected in correspondence to the domains of datasets we have tested in our work.

\begin{figure*}[!t]
    \centering
    \includegraphics[width=0.8\textwidth]{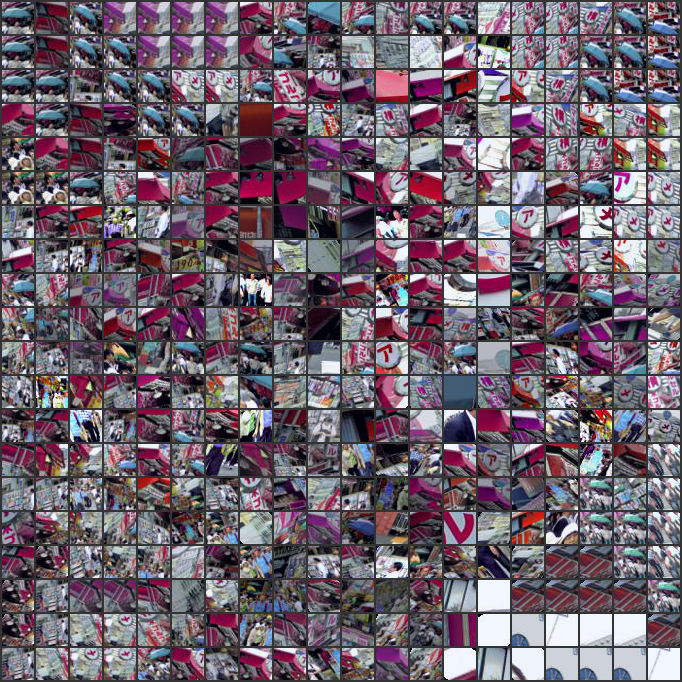}
    \caption{t-SNE Manifold visualization of distillation dataset corresponding to t-SNE scatter plot in Fig. \ref{fig:img_viz:tsne_high_accuracy}).}
    \label{fig:img_viz:manifold_viz}
\end{figure*}

\subsection{Visualisation of Image Representations from Single Images}\label{App:img_viz:crops}

In Fig. \ref{fig:img_viz:tsne_scatterplot}, we visualize the TSNE embeddings of the training examples in our single image distillation dataset during different phases of training on a scatterplot. Its implications have been discussed in the main paper in brief detail. To add to the earlier analysis, by looking at the difference between Fig. \ref{fig:img_viz:tsne_low_accuracy} and Fig. \ref{fig:img_viz:tsne_high_accuracy}, one can see how the embedding layer in the neural network starts forming particular regions belonging to certain classes with the increase in classification accuracy. Even though the image representations do not contain many of the classes in the classifier as is prevalent from visual inspection of Fig. \ref{fig:img_viz:manifold_viz}, the pseudo labels are powerful enough to enable the learning through knowledge distillation in a federated setting.

\section {Finding Optimal Hyper-Parameters for Selection Algorithms}\label{App:optimal_selection}

\subsection{KMeans Balancing}\label{App:optimal_selection:kmeans}

To examine the impact of KMeans Balancing and its corresponding settings on the federated training, we conduct experiments while varying their values on multiple datasets having different numbers of prediction labels. The results have been presented in Table \ref{tab:abl_kmeans}. Although values do not differ by a large margin for CIFAR10, CIFAR100 results provide us more assurance for the optimal values. We find that the KMeans selection strategy works best with a high number of clusters (K) compared to the number of classes in the corresponding classification problem (Table \ref{tab:abl_kmeans:number}), while forcing maximum class balancing while pruning (Table \ref{tab:abl_kmeans:balancing_factor}) as well as selecting the hard examples (Table \ref{tab:abl_kmeans:heuristics}) for distillation. Since the KMeans model is an independent model working on giving the pseudo-labels to the distillation training examples, there could be a play of correlation between the 3 hyper-parameters for this mechanism. It can be an interesting point for future research on this.

\begin{table}[htbp]
    \centering
    \begin{subtable}[b]{0.47\textwidth}
        {\small
\centering
\begin{tabular}{lll}
    \toprule
    \multicolumn{1}{c}{\multirow{2}{*}{\textbf{Number of Clusters}}} & \multicolumn{2}{c}{\textbf{Private Dataset}} \\
    \cmidrule(lr){2-3}
     & CIFAR10 & CIFAR100 \\
    \midrule
    5   & 64.3 $\pm$ 0.7 & 19.9 $\pm$ 1.9 \\
    10  & 65.2 $\pm$ 0.9 & 21.2 $\pm$ 0.1 \\
    50 & 65.3 $\pm$ 0.3 & 22.9 $\pm$ 0.9 \\
    100 & 65.5 $\pm$ 1.1 & 18.9 $\pm$ 2.9 \\
    1000 & \textbf{66.3 $\pm$ 1.1} & \textbf{22.9 $\pm$ 0.1} \\
    \bottomrule
\end{tabular}%
\caption{Varying cluster number with \textbf{selection heuristic = easy} and \textbf{balancing factor = 0.1}.}
\label{tab:abl_kmeans:number}}
    \end{subtable}
    \hfill
    \begin{subtable}[b]{0.47\textwidth}
        {\small
\centering
\begin{tabular}{lll}
    \toprule
    \multicolumn{1}{c}{\multirow{2}{*}{\textbf{Balancing Factor}}} & \multicolumn{2}{c}{\textbf{Private Dataset}} \\
    \cmidrule(lr){2-3}
     & CIFAR10 & CIFAR100 \\
    \midrule
    0.0 & 64.9 $\pm$ 1.3 & 22.5 $\pm$ 0.2 \\
    0.05  & 65.0 $\pm$ 1.7  & 22.5 $\pm$ 0.4 \\
    0.1 & 65.2 $\pm$ 2.0 & 21.3 $\pm$ 0.1 \\
    0.5 & \textbf{66.6 $\pm$ 1.4} & 22.2 $\pm$ 0.1 \\
    1.0 & 66.0 $\pm$ 1.2 & \textbf{24.0 $\pm$ 1.4} \\
    \bottomrule
\end{tabular}%
\caption{Varying balancing factor with \textbf{number of clusters (K) = 1000} and \textbf{selection heuristic = easy}.}
\label{tab:abl_kmeans:balancing_factor}}
    \end{subtable}
    \newline \newline
    \begin{subtable}[b]{0.47\textwidth}
        {\small
\centering
\begin{tabular}{lll}
    \toprule
    \multicolumn{1}{c}{\multirow{2}{*}{\textbf{Selection Heuristics}}} & \multicolumn{2}{c}{\textbf{Private Dataset}} \\
    \cmidrule(lr){2-3}
     & CIFAR10 & CIFAR100 \\
    \midrule
    Easy & 65.8 $\pm$ 1.3 & 22.1 $\pm$ 0.4\\
    Hard  & \textbf{66.7 $\pm$ 1.1} & \textbf{23.8 $\pm$ 1.8} \\
    Mixed (50-50) & 65.5 $\pm$ 2.1 & 21.8 $\pm$ 1.0 \\
    \bottomrule
\end{tabular}%
\caption{Varying selection heuristics with \textbf{number of clusters (K) = 1000} and \textbf{balancing factor = 1.0}.}
\label{tab:abl_kmeans:heuristics}}
    \end{subtable}
    
    \caption{Best test set accuracy achieved during 30 rounds of training with KMeans balancing under different settings with different private datasets (Distribution $\alpha = 1.0$) using single image patches as the distillation dataset with 20\% FedAvg initialisation rate on ResNet-8 (across 2 seeds).}
    \label{tab:abl_kmeans}
\end{table}

\subsection{Entropy Selection}\label{App:optimal_selection:entropy}

For the entropy selection mechanism, we only have 2 settings to vary: Removal Percentage and Heuristics (Section \ref{ss:method:select:entropy}). The results obtained during ablation studies with these settings have been presented in Table \ref{tab:abl_entropy}. From the results in Table \ref{tab:abl_entropy:heuristic}, it is clear that removing training examples with high confidence (less entropy) provides us best results. Removing a high percentage of training examples from a large initial set using this mechanism also provided us with more robust training, compared to removing a smaller number of training examples from a small initial set (Table \ref{tab:abl_entropy:k}). Similar to our last experiments with the KMeans mechanism, the results are more clearly pronounced in the presence of a more difficult dataset (100 classes compared to 10 in the case of CIFAR100 and CIFAR10).

\begin{table}[htbp]
    \centering
    \begin{subtable}[b]{0.47\textwidth}
        {\small
\centering
\begin{tabular}{lll}
    \toprule
    \multicolumn{1}{c}{\multirow{2}{*}{\textbf{Removal Heuristic}}} & \multicolumn{2}{c}{\textbf{Private. Dataset}}\\
    \cmidrule(lr){2-3}
     & \multicolumn{1}{c}{CIFAR10} & \multicolumn{1}{c}{CIFAR100}\\
    \midrule
    Top & 67.0 $\pm$ 0.9 & \textbf{26.2 $\pm$ 1.9}\\
    Bottom & 61.9 $\pm$ 2.4 & 15.4 $\pm$ 2.4\\
    Random & \textbf{67.4 $\pm$ 1.5} & 23.2 $\pm$ 0.9\\
    \bottomrule
\end{tabular}%
\caption{Varying Removal Heuristic with \textbf{Removal Percentage = 90\%}.}
\label{tab:abl_entropy:heuristic}}
    \end{subtable}
    \hfill
    \begin{subtable}[b]{0.47\textwidth}
        {\centering
\small
\begin{tabular}{lll}
    \toprule
    \multicolumn{1}{c}{\multirow{2}{*}{\textbf{Removal Percentage (\%)}}} & \multicolumn{2}{c}{\textbf{Private Dataset}}\\
    \cmidrule(lr){2-3}
     & \multicolumn{1}{c}{CIFAR10} & \multicolumn{1}{c}{CIFAR100}\\
    \midrule
    10 & 65.0 $\pm$ 0.9 & 24.4 $\pm$ 1.2\\
    50 & \textbf{67.5 $\pm$ 1.2} & 25.8 $\pm$ 1.4\\
    90 & 66.2 $\pm$ 1.5 & \textbf{26.7 $\pm$ 1.5}\\
    \bottomrule
\end{tabular}%
\caption{Varying Removal Percentage with \textbf{Removal Heuristic = Top}.}
\label{tab:abl_entropy:k}}
    \end{subtable}
    \caption{Best test set accuracy achieved during 30 rounds of training with Entropy selection under different settings with different private datasets (Distribution $\alpha = 1.0$) using single image patches as the distillation dataset with 20\% FedAvg initialisation rate on ResNet-8 (across 2 seeds).}
    \label{tab:abl_entropy}
\end{table}

\section{Learning Rate Optimisation}\label{App:learn_rate}

\subsection{Client Side LR}\label{App:learn_rate:client}

We find the optimal learning rate for local model training by using vanilla FedAvg as represented in Table \ref{tab:client_lr} on ResNet-8. We do a grid search across a certain set of values for it, which is described in Table \ref{tab:client_lr}. 

\begin{table}[htbp]
    \small
    \centering
    \begin{tabular}{l|l}
    \toprule
    \multicolumn{1}{c|}{\textbf{Local L.R.}} & \multicolumn{1}{c}{\textbf{Accuracy}} \\
    \midrule
    0.1   & 74.6 \\
    0.05  & 80.1 \\
    0.01 & \textbf{80.9} \\
    0.005 & 80.6 \\
    0.001 & 74.6 \\
    \bottomrule
    \end{tabular}
    \caption{Highest test accuracy achieved during 30 rounds of training with FedAvg (no distillation) on CIF10 with ResNet-8 (Distribution $\alpha=100.0$).}
    \label{tab:client_lr}
\end{table}

\subsection{Server Side LR}\label{App:learn_rate:global}

To find the optimal learning rate for the distillation training, we use FedDF as depicted in Table \ref{tab:server_lr}. This is accomplished with the help of a grid search across certain values for the distillation training learning rate, as mentioned in Table \ref{tab:server_lr} itself.

\begin{table}[htbp]
    \small
    \centering
    \begin{tabular}{l|l}
    \toprule
    \multicolumn{1}{c|}{\textbf{Global L.R.}} & \multicolumn{1}{c}{\textbf{Accuracy}} \\
    \midrule
    0.1   & 44.8 $\pm$ 1.4 \\
    0.05  & 62.7 $\pm$ 1.0 \\
    0.01 & 76.6 $\pm$ 0.3 \\
    0.005 & \textbf{78.1 $\pm$ 0.4} \\
    0.001 & 75.2 $\pm$ 1.0 \\
    \bottomrule
    \end{tabular}
    \caption{Highest test accuracy achieved during 30 rounds of training with client learning rate = 0.01 using FedDF on CIF10 with ResNet-8 on 2 different seeds. (Distribution $\alpha = 100.0$)}
    \label{tab:server_lr}
\end{table}

\end{document}